\documentclass{article}
\usepackage{ijcai26}

\usepackage{times}
\usepackage{soul}
\usepackage{url}
\usepackage[hidelinks]{hyperref}
\usepackage[utf8]{inputenc}
\usepackage[small]{caption}
\usepackage{graphicx}
\usepackage{amsmath}
\usepackage{amsthm}
\usepackage{booktabs}
\usepackage{algorithm}
\usepackage{algorithmic}
\usepackage[switch]{lineno}

\usepackage{multirow} 
\usepackage{amssymb}
\usepackage{amsfonts}
\usepackage{pifont}
\usepackage{listings}

\usepackage{multirow}
\usepackage[table]{xcolor}

\title{Harmonizing Real-Time Constraints and Long-Horizon Reasoning: An Asynchronous Agentic Framework for Dynamic Scheduling}

\author{
Shijie Cao$^{1,2}$
\and
Yuan Yuan$^{1,3,4}$\footnote{Corresponding author.}\And
Jing Liu$^{2,5,6}$\\
\affiliations
$^1$School of Computer Science and Engineering, Beihang University, Beijing 100191, China\\
$^2$Shenzhen Loop Area Institute, Shenzhen, China\\
$^3$Qingdao Research Institute, Beihang University\\
$^4$Hangzhou Innovation Institute, Beihang University\\
$^5$School of Artificial Intelligence, Xidian University, Xi'an 710071, Shaanxi, China \\
$^6$Guangzhou Institute of Technology, Xidian University, Guangzhou 510555, Guangdong, China \\
\emails
\{cls1277, yuan21\}@buaa.edu.cn,
neouma@mail.xidian.edu.cn
}

\begin{document}

\maketitle

\begin{abstract}
The Dynamic Flexible Job Shop Scheduling Problem (DFJSP) necessitates a trade-off between instant reaction to stochastic disturbances and global optimization of production goals. Conventional priority rules are insufficiently flexible to handle complex disruptions, whereas learning-based approaches often compromise interpretability or fail to generalize across problem scales. Although Large Language Models (LLMs) offer advanced reasoning capabilities to bridge this gap, their substantial inference latency is incompatible with the millisecond-level decision cycles of industrial control systems. To resolve this conflict, we introduce RACE-Sched, an asynchronous agent-based framework that decouples policy execution from logical reasoning via a dual-stream architecture. The Reactive Stream executes low-latency symbolic heuristics to enable real-time dispatching, while the parallel Deliberative Stream leverages an LLM to synthesize, validate, and evolve these rules. Candidate rules undergo rigorous testing in a sandbox and are deployed via atomic updates, ensuring safety without blocking the control loop. Additionally, a semantic rule repository indexes validated heuristics for retrieval-based initialization which enhances transferability across problem scales. Extensive evaluations on GEN-Bench, MK-Bench, and JMS-Bench demonstrate that RACE-Sched outperforms leading Deep Reinforcement Learning and other LLM-based baselines. This approach harmonizes real-time constraints with long-horizon reasoning to achieve superior solution quality and robust adaptation to dynamic events.
\end{abstract}

\section{Introduction}

The Dynamic Flexible Job Shop Scheduling Problem (DFJSP) is a core online decision problem in modern manufacturing, where each ready operation must be assigned to a feasible heterogeneous machine under routing flexibility and precedence constraints~\cite{xu2025learn}.
When machine failures, urgent job arrivals, or processing-time variations change the shop-floor state, delayed dispatching decisions can block available machines and propagate idle time through subsequent operations.

Data-driven dispatching policy learning is achievable with Deep Reinforcement Learning (DRL), but interpreting its decisions poses significant challenges and these decisions may be vulnerable to variations in training details~\cite{zhang2020learning}.
Priority Dispatching Rules (PDRs) are easy to execute and inspect, yet their rigid structures often fail to adapt when the bottleneck shifts~\cite{holthaus2000efficient}.
By generating executable rules with LLMs, Code as Policy (CaP) improves the transparency of scheduling decisions~\cite{liang2022code}.
One remaining obstacle is response latency: making LLM calls during the control loop operation is too slow to satisfy the real-time demands of scheduling systems.

The fundamental conflict lies in the temporal granularity between reasoning and execution. Advanced reasoning models such as GPT-4 require several seconds to process context and generate code~\cite{openai2024gpt4ocard}. In contrast, dynamic scheduling decisions in a high-speed production line must be made within milliseconds to avoid blocking machine operations~\cite{singh2026areview,ouelhadj2009asurvey}. Simple integration strategies that pause the production line to await external reasoning result in unacceptable throughput losses. This leads to a frequency mismatch where the slow cognitive cycle of the LLM cannot synchronize with the fast physical cycle of the manufacturing floor~\cite{karami2025exploringdynamicschedulingspace}. Consequently, current approaches are forced to compromise by either relying on pre-generated static rules that lack adaptability or employing smaller and less capable models that sacrifice reasoning quality for speed~\cite{ankit2026smalllm}.

Furthermore, existing hybrid frameworks typically separate learning from execution in a manner that precludes continuous online adaptation. Most methodologies rely on an offline training phase, where scheduling rules are derived from historical data~\cite{priore2014dynamic}. Once deployed, these rules remain static and cannot respond to unseen disturbances such as sudden machine breakdowns or drastic shifts in order priority. Although some adaptive systems attempt to periodically retrain policies, they often require synchronous interactions that disrupt the ongoing scheduling process. A key limitation in current research is the lack of mechanisms that allow for the safe and asynchronous evolution of control logic during system operations.
An ideal scheduling system should enable the control logic to improve continuously without impeding the real-time responsiveness of physical machines.

While inspired by the asynchronous dual-stream concept~\cite{chen2025fastinslowdualsystemfoundationmodel}, directly applying existing embodied agents to industrial scheduling is hindered by the requirement for interpretability and strict safety guarantees.
Rather than treating LLM-based code generation, simulation validation, and iterative refinement as new ingredients, we focus on how these mechanisms can be safely integrated into a running DFJSP control loop.
We introduce RACE-Sched\footnote{To facilitate reproducibility, our code is available at \url{https://github.com/cls1277/RACE-Sched}}, a framework where the Reactive Stream runs low-latency symbolic heuristics, while the Deliberative Stream leverages an LLM to analyze summary statistics from a sliding window of recent decisions, generate candidate Python heuristics, and test them in a sandbox that replays representative instances under the same constraints.
Only candidates that meet predefined acceptance criteria are promoted.
The Reactive Stream does not lie on the critical path: it never waits for deliberation, and updates are executed through an atomic pointer swap of the active rule set.

We also maintain a rule repository that indexes validated heuristics together with lightweight instance-level metadata, such as numbers of jobs and machines. 
The Deliberative Stream retrieves the most similar rules to warm-start candidate generation and sandbox evaluation, which helps shorten prompts and enhance robustness across problem scales.

Our contributions are as follows.
\begin{itemize}
    \item An asynchronous symbolic evolution framework adapted for industrial constraints, which decouples real-time dispatching from LLM-driven heuristic synthesis via a safe ``code-as-policy'' mechanism, ensuring the control loop remains responsive.
    \item A sandbox-based validation and safe deployment mechanism, where candidate heuristics are evaluated offline through constrained replay, and promoted only if they meet predefined acceptance criteria, followed by an atomic pointer swap of the active rule set.
    \item A rule repository that indexes validated heuristics with lightweight instance-level metadata, enabling warm-start retrieval and improving transferability across different benchmarks and problem scales.
    \item An extensive evaluation on GEN-Bench, MK-Bench, and JMS-Bench, including a machine-failure stress test, showing that our approach achieves higher solution quality and faster adaptation than competitive DRL baselines and direct LLM control.
\end{itemize}

\section{Related Work}

\paragraph{Heuristics and Evolutionary Scheduling.}
Priority Dispatching Rules remain the industrial standard for dynamic scheduling due to their minimal computational cost of constant time complexity and ease of interpretation~\cite{holthaus2000efficient}. However, fixed rules such as Shortest Processing Time and Most Work Remaining are inherently myopic. They rely on local buffer status and often fail to maintain performance when system bottlenecks shift due to stochastic disturbances. To automate the design of dispatching policies, Genetic Programming (GP) has been widely applied to evolve symbolic priority functions that combine production attributes into complex rules~\cite{mei2016feature}. While advances such as surrogate-assisted GP~\cite{mei2016feature} and multitask evolution~\cite{zhang2022multitask} have improved convergence speed, the evolutionary process remains computationally expensive and offline. This limitation prevents GP from achieving the rapid online adaptation required to handle sudden machine failures or urgent order insertions in real-time environments.

\paragraph{DRL for Dynamic Scheduling.}
To capture complex system dynamics, recent research has shifted toward DRL by modeling scheduling as a Markov Decision Process~\cite{xu2025learn}. State representation has evolved from simple feature vectors to Graph Neural Networks that encode the non-Euclidean topology of job shops~\cite{zhang2025meta}. The Dual Attention Network~\cite{wang2023flexible} established a benchmark by jointly attending to operation precedence and machine contention. Specialized architectures have also been developed for specific disruptions such as IDDQN~\cite{wu2025dynamic} for robustness against machine breakdowns and hierarchical frameworks such as HMPSAC~\cite{ding2025data} for multi-objective trade-offs. Despite these gains, DRL faces critical deployment barriers. The resulting black-box neural policies lack the interpretability required for safety-critical manufacturing~\cite{li2026multiobj} and often exhibit poor generalization when transferring between different problem scales.

\begin{figure*}[!t]
\centering
\includegraphics[width=\textwidth]{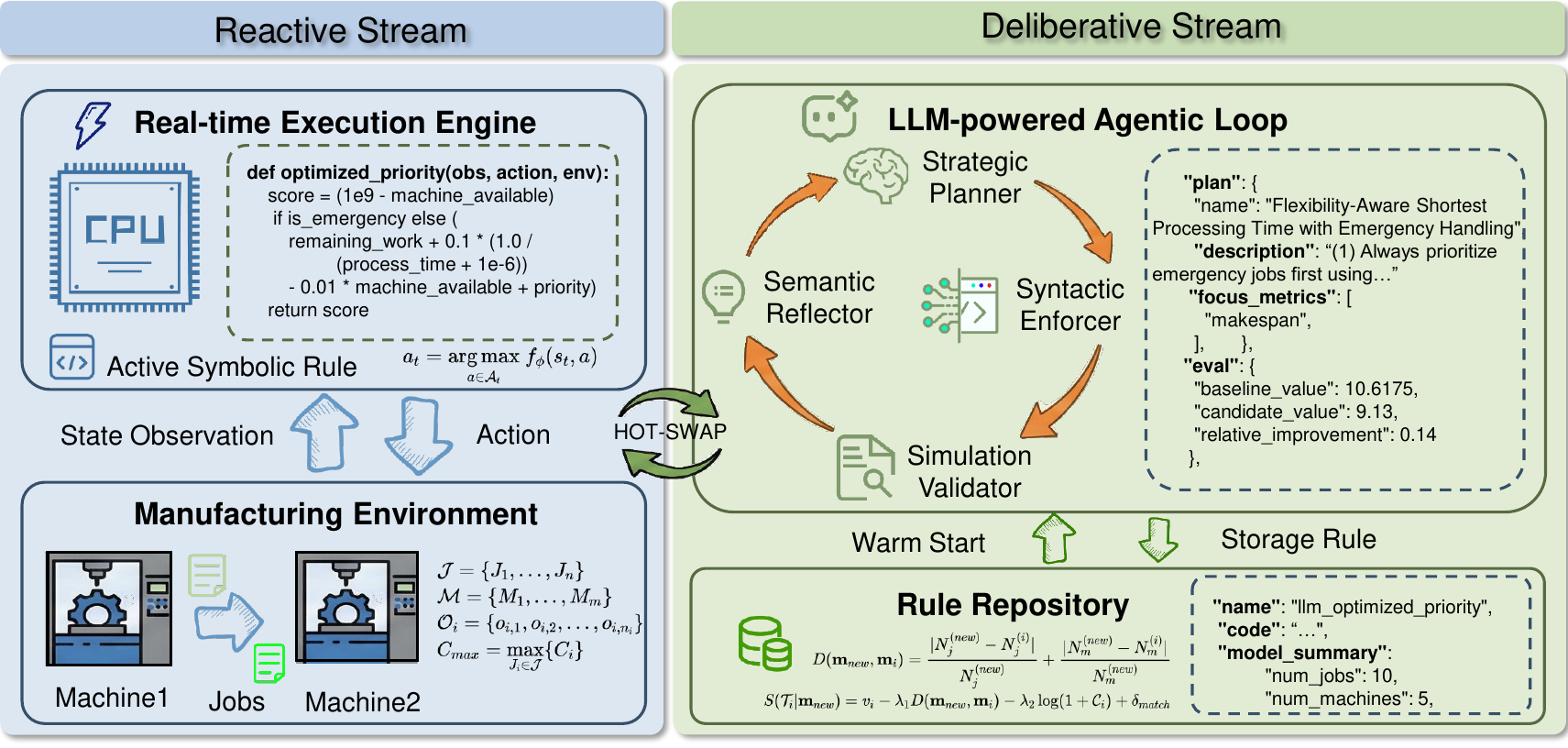}
\caption{Overview of RACE-Sched. The Reactive Stream executes an active symbolic rule for real-time scheduling, while the Deliberative Stream runs an LLM-driven loop with constrained code generation and sandbox evaluation to produce validated rule updates. Validated heuristics are stored in a rule repository for warm-start retrieval and deployed to the control loop via hot-swap.}
\label{fig:framework_arch}
\end{figure*}

\paragraph{LLM Reasoning and Code-as-Policy.}
LLMs offer a promising avenue to address the interpretability and reasoning gaps in DRL. While techniques such as Chain-of-Thought~\cite{wei2022chain} and Tree of Thoughts~\cite{yao2023tree} enhance logical planning, the high latency of inference creates a temporal mismatch with the millisecond-level response requirements of industrial control. To address this discrepancy, the CaP paradigm proposes generating executable programs rather than direct text actions~\cite{liang2022code}. This approach generates fast and interpretable code as validated by open-ended agents such as Voyager~\cite{wang2023voyager}. In the scheduling domain, ReflecSched~\cite{cao2025reflecsched} recently utilized LLMs to generate hierarchical reflections for guidance.
Prior LLM reflection and CaP scheduling methods mainly focus on synthesizing or refining executable policies, while hybrid offline-online heuristic evolution and safe update schemes primarily address policy improvement and deployment risk as separate stages. RACE-Sched instead focuses on the control-integration problem: it confines slow LLM deliberation, sandbox validation, and candidate policy improvement to the background, while the online scheduler executes only the currently validated symbolic rule and receives accepted updates through atomic hot-swap.

\section{Problem Formulation and Preliminaries}

We consider the DFJSP, which entails the assignment of $n$ jobs $\mathcal{J} = \{J_1, \dots, J_n\}$ to a set of $m$ heterogeneous machines $\mathcal{M} = \{M_1, \dots, M_m\}$.
Each job $J_i \in \mathcal{J}$ comprises a strictly ordered sequence of $n_i$ operations $\mathcal{O}_i = \{o_{i,1}, o_{i,2}, \dots, o_{i,n_i}\}$.
Routing flexibility permits each operation $o_{i,j}$ to be executed on any machine within an eligible subset $\mathcal{M}_{i,j} \subseteq \mathcal{M}$, where $p_{i,j,k}$ denotes the deterministic processing time on machine $M_k \in \mathcal{M}_{i,j}$~\cite{cao2023inverse}.

The system state at time $\tau$ comprises the progress of active jobs, machine availability, and the backlog of pending operations.
Temporal evolution is driven by asynchronous stochastic disturbances that induce discrete state transitions, including non-deterministic job arrivals and machine failure-recovery cycles.
Under a preempt-resume scheduling policy, interrupted operations remain suspended and are resumed from the point of interruption upon the restoration of machine capacity.
The objective is to minimize the expected makespan $C_{\max}$, where $C_i$ denotes the completion time of job $J_i$ and $C_{\max}=\max_{J_i\in\mathcal{J}} C_i$, subject to real-time decision constraints in a dynamic manufacturing environment.

We formulate this task as a sequential decision-making process where a policy $\pi$ maps system states to operation-machine assignments at each decision epoch $t$, while satisfying precedence and availability constraints.
In this stochastic setting, performance optimization requires strategic reasoning to address complex operational tradeoffs, such as the proactive allocation of machine capacity to alleviate anticipated bottlenecks.

\section{Methodology}
\label{sec:methodology}

A detailed workflow diagram of RACE-Sched is provided in Figure~\ref{fig:framework_arch}. The Reactive Stream runs a symbolic priority rule inside the control loop. 
A rule repository is maintained to store validated heuristics together with lightweight instance-level metadata, facilitating retrieval-based warm starts.
The Deliberative Stream improves the active rule asynchronously using compact summaries of recent states and example actions, together with constrained code generation and sandbox evaluation procedures.

\subsection{Asynchronous Dual Stream Architecture}

We model DFJSP as a Semi-Markov Decision Process with state $s_t$ and action $a_t$ at each decision epoch $t$~\cite{chang2022deeprl}. The Reactive Stream maps $s_t$ to a dispatching action using a symbolic rule, while the Deliberative Stream performs rule updates using LLM-based reasoning, which operates on a slower timescale.

\subsubsection{Reactive Stream for Execution}

The Reactive Stream runs inside the scheduling loop, with the objective of confining the decision latency within the millisecond range~\cite{fristiane2022effective}. It executes a symbolic priority function $f_\phi$ implemented as a compiled Python rule derived from the current code representation. At decision epoch $t$, it selects
\begin{equation}
    a_t = \underset{a \in \mathcal{A}_t}{\arg\max} \ f_\phi(s_t, a)
    \label{eq:priority_rule}
\end{equation}
The interpreter is constrained to use only observable features in $s_t$ and does not permit side effects, thereby the rule remains deterministic and operates with millisecond-scale execution time.

\subsubsection{Deliberative Reasoning Stream for Policy Evolution}

The Deliberative Stream updates the symbolic rule off the critical path.
It takes a compact summary of recent states and example actions together with the current rule $\phi_{\text{old}}$ and retrieved knowledge $\mathcal{K}$ and proposes a new rule~\cite{yang2023large,gao2025surveyselfevolvingagentspath}
\begin{equation}
    \phi_{\text{new}} \leftarrow \mathcal{P}_{\text{delib}}(\xi, \phi_{\text{old}}, \mathcal{K})
\end{equation}
where $\xi$ denotes the summarized observation window.
The Reactive Stream keeps running $\phi_{\text{old}}$ while the proposal is generated and tested in the sandbox.

\subsubsection{Trigger Update Protocol and Atomic Transition}

Deliberation is triggered either every $K$ decision steps or when a monitored metric drops relative to its sliding-window best of length $L$.
Let $m_t$ denote the metric and $m_t^{\ast}=\max_{i\in[t-L+1,t]} m_i$.
Define the periodic trigger indicator as
\begin{equation}
\mathbb{I}_{\text{periodic}}(t) = \mathbb{I}\{t \bmod K = 0\},
\end{equation}
and the performance-based trigger indicator as
\begin{equation}
\mathbb{I}_{\text{perf}}(t) = \mathbb{I}\left\{\frac{m_t^{\ast} - m_t}{|m_t^{\ast}|} \ge \epsilon \right\}.
\end{equation}
When $\mathbb{I}_{\text{periodic}}(t)\lor\mathbb{I}_{\text{perf}}(t)$ holds, the Deliberative Stream proposes $\phi_{\text{new}}$ and atomically swaps it in only if sandbox screening accepts it.

\subsection{The Agentic Loop: A Safety-Constrained Evolutionary Cycle}

The Strategic Planner produces a compact directive based on the current system profile. 
The Syntactic Enforcer converts this directive into an executable priority dispatching rule, subject to syntactic safety and complexity constraints. 
The Simulation Validator evaluates candidate rules in a sandbox, allowing an update only if it meets predefined feasibility and improvement criteria. Concurrently, the Semantic Reflector records concise diagnostics to justify the acceptance or rejection decision for human inspection.

\subsubsection{Strategic Planner for Bottleneck-Aware Reasoning}

The Strategic Planner generates a compact profile by summarizing the current state and formulates a local strategy, such as shifting priority toward the current bottleneck or favoring short remaining processing time on a congested machine. We formalize this mechanism as
\begin{equation}
    \mathcal{S}_{plan} \leftarrow \operatorname{Planner}(s_{profile}, \mathcal{M}_{obj})
\end{equation}
where $\mathcal{M}_{obj}$ is the objective and $\mathcal{S}_{plan}$ is a short directive consumed by the Syntactic Enforcer.

\subsubsection{Syntactic Enforcer for Executable Logic Generation}

The Syntactic Enforcer converts the strategic plan $\mathcal{S}_{plan}$ into executable logic in the form of PDRs.
It produces symbolic Python functions whose source code can be inspected by domain experts.

Rule generation relies on constrained prompting, which encourages the adoption of a standardized set of shop floor attributes and per operation features.
The Coder outputs a priority function $f_\phi$ that scores each feasible operation and induces a dispatching decision.

The Syntactic Enforcer enforces basic safety and executability by compiling candidate code in a restricted execution context with a limited set of built-in functions and restricted import permissions.
If the generated code exceeds a fixed length threshold, it is truncated before compilation, and non-compilable candidates are discarded.
The module further extracts lightweight complexity statistics for bookkeeping and subsequent downstream selection.

\subsubsection{Simulation Validator for Sandbox Screening}

The Simulation Validator screens a candidate policy $\phi_{cand}$ against the incumbent $\phi_{old}$ in a sandbox before deployment~\cite{liu2023isyourcode}.
On an evaluation pool $\mathcal{D}_{val}$, it runs rollouts for both policies under the same decision constraints and disturbance process, yielding per episode objective values $\{y_i^{old}\}_{i=1}^{n}$ and $\{y_i^{cand}\}_{i=1}^{n}$.
Define the empirical estimate
\begin{equation}
\widehat{\mathcal{M}}_{obj}(\phi)=\frac{1}{n}\sum_{i=1}^{n}y_i^{\phi}\, .
\end{equation}
For a minimization objective, the relative improvement is
\begin{equation}
r_t=\frac{\widehat{\mathcal{M}}_{obj}(\phi_{old})-\widehat{\mathcal{M}}_{obj}(\phi_{cand})}{d_t}\, ,
\end{equation}
with
\begin{equation}
d_t=
\begin{cases}
\left|\widehat{\mathcal{M}}_{obj}(\phi_{old})\right|, & \left|\widehat{\mathcal{M}}_{obj}(\phi_{old})\right|>10^{-9}\\
1.0, & \text{otherwise.}
\end{cases}
\end{equation}
Let $\mu_{old},\mu_{cand}$ and $s_{old}^2,s_{cand}^2$ be the sample means and variances of the two rollout sets.
The validator computes
\begin{equation}
s_p^2=\frac{(n-1)s_{old}^2+(n-1)s_{cand}^2}{2n-2}\, ,
\end{equation}
\begin{equation}
d_t^{eff}=\frac{\mu_{old}-\mu_{cand}}{\sqrt{s_p^2}}\, ,
\end{equation}
and
\begin{equation}
T_t=\frac{\mu_{old}-\mu_{cand}}{\sqrt{s_{old}^2/n+s_{cand}^2/n}}\, .
\end{equation}
A candidate is accepted only if sandbox execution is feasible and
\begin{equation}
\mathbb{I}_{acc}=\mathbb{I}\left\{
\begin{aligned}
&r_t\ge\epsilon_{rel}\;\land\; d_t^{eff}\ge d_{min}\\
&\land\; T_t\ge T_{\alpha}
\end{aligned}
\right\}\, ,
\end{equation}
where $\epsilon_{rel}$, $d_{min}$, and $T_{\alpha}$ are fixed thresholds.
The validator returns $\mathcal{E}_{cand}$ containing $\widehat{\mathcal{M}}_{obj}(\phi_{old})$, $\widehat{\mathcal{M}}_{obj}(\phi_{cand})$, $r_t$, $d_t^{eff}$, the number of episodes used, and $\mathbb{I}_{acc}$.

\subsubsection{Semantic Reflector for Feedback and Validation}

The Semantic Reflector completes the optimization loop by converting sandbox evaluation data into actionable feedback for the next iteration.
Functioning as an interpreter, it synthesizes observed tradeoffs and provides a concise diagnostic assessment that rationalizes the acceptance or rejection of each candidate.

This diagnostic process maps sandbox evaluation statistics and high-level policy descriptions into structured feedback.
In cases where a candidate policy improves makespan but degrades maximum tardiness, the module interprets the statistical summaries under the chosen objective, produces concise comments that highlight the tradeoff, and determines whether further exploration is warranted.

Diagnostic feedback is logged in the trajectory records, where it serves to rationalize the acceptance or rejection of individual candidates.
Its decision signal controls whether the search runs another iteration. The final rule selection, meanwhile, depends solely on sandbox evaluation results and a complexity term.

\subsection{Rule Repository for Lifelong Learning}

To mitigate knowledge volatility inherent in conventional reinforcement learning and enhance adaptation efficiency across heterogeneous environments, we introduce the Semantic Knowledge Repository. This module persists and retrieves optimized symbolic heuristics, serving as reusable knowledge for RACE-Sched. 
Through the decoupling of knowledge retention from parameter updates, it preserves high-performing heuristics under nonstationary dynamics. 
During runtime, the retrieval of validated priors enables faster adaptation than re-optimizing from scratch.

\subsubsection{Formal Structure of the Knowledge Repository}

The Semantic Knowledge Repository is organized as an indexed library of interpretable symbolic heuristics. Each entry constitutes a tuple $\mathcal{T}_i = \langle \mathbf{m}_i, \phi_i, v_i, \mathcal{C}_i \rangle$ that links a synthesized policy to its operational context. The meta feature vector $\mathbf{m}_i$ encodes global problem topology including job and machine cardinality to facilitate similarity based retrieval within the vector space. The component $\phi_i$ represents the executable policy source code while the scalar $v_i$ quantifies the normalized performance improvement over the baseline. To promote generalizability we incorporate a structural complexity score $\mathcal{C}_i$ derived from code statistics such as operator count and branching factor. This metric functions as a regularization penalty during retrieval and prioritizes parsimonious logic that avoids overfitting to specific training instances.

\subsubsection{Retrieval and Cross Scenario Transfer}

Upon encountering a new problem instance $\mathbf{m}_{new}$, the system initiates a warm start by retrieving the nearest policy neighbors from the repository.
The retrieval score $S(\mathcal{T}_i | \mathbf{m}_{new})$ balances historical efficacy, feature alignment, and complexity:
\begin{equation}
    S(\mathcal{T}_i | \mathbf{m}_{new}) = v_i - \lambda_1 D(\mathbf{m}_{new}, \mathbf{m}_i) - \lambda_2 \log(1 + \mathcal{C}_i) + \delta_{match}
\end{equation}
where $v_i$ is the policy efficacy, $D(\mathbf{m}_{new}, \mathbf{m}_i)$ is the feature distance between the new instance $\mathbf{m}_{new}$ and repository entry $\mathbf{m}_i$, and $\mathcal{C}_i$ is the policy complexity.

The distance $D(\mathbf{m}_{new}, \mathbf{m}_i)$ is calculated as:
\begin{equation}
    D(\mathbf{m}_{new}, \mathbf{m}_i) = \frac{|N_j^{(new)} - N_j^{(i)}|}{N_j^{(new)}} + \frac{|N_m^{(new)} - N_m^{(i)}|}{N_m^{(new)}}
\end{equation}
where $N_j$ and $N_m$ represent job volume and machine capacity, respectively.

This regularization prevents maladaptive transfer by penalizing shifts in job volume and machine capacity.

\section{Experiments}

\subsection{Experimental Setup}

\begin{table*}[t]
\centering
\begin{tabular}{llrrrrrrrr}
\toprule
\textbf{Category} & \textbf{Model Backend} & \multicolumn{2}{c}{\textbf{GEN-Bench Small}} & \multicolumn{2}{c}{\textbf{GEN-Bench Normal}} & \multicolumn{2}{c}{\textbf{MK-Bench}} & \multicolumn{2}{c}{\textbf{JMS-Bench}} \\
\cmidrule(lr){3-4} \cmidrule(lr){5-6} \cmidrule(lr){7-8} \cmidrule(lr){9-10}
& & RPD (\%) & Rank & RPD (\%) & Rank & RPD (\%) & Rank & RPD (\%) & Rank \\
\midrule
\textbf{Baselines} & GP & 39.20 & 20.50 & 72.41 & 23.00 & 62.88 & 17.20 & 51.25 & 18.30 \\
& HMPSAC & 12.03 & 12.06 & 12.34 & 14.55 & 20.17 & 11.80 & 12.87 & 9.20 \\
& IDDQN & 13.58 & 14.11 & 9.82 & 12.10 & 17.47 & 11.20 & 11.02 & 8.30 \\
& DAN & 13.47 & 13.44 & 10.40 & 12.55 & 23.74 & 14.00 & 18.88 & 13.60 \\
& PPO-OC & 17.33 & 15.72 & 12.23 & 15.25 & 21.60 & 11.60 & 14.13 & 10.60 \\
\midrule
\textbf{LLM-Direct} & Qwen3-8B & 11.13 & 14.50 & 10.15 & 13.00 & 29.62 & 15.30 & 15.77 & 12.40 \\
& Qwen3-14B & 7.86 & 8.39 & 10.48 & 13.55 & 13.44 & 7.80 & 16.57 & 12.20 \\
& Qwen3-32B & 9.47 & 11.17 & 10.70 & 13.30 & 23.58 & 13.40 & 11.98 & 8.60 \\
\midrule
\textbf{ReflecSched} & Qwen3-8B & 8.73 & 10.56 & 7.44 & 9.60 & 15.94 & 9.10 & 9.56 & 7.70 \\
& Qwen3-14B & 6.71 & 7.11 & 7.67 & 9.55 & 20.20 & 11.60 & 9.29 & 7.50 \\
& Qwen3-32B & 7.10 & 8.50 & 8.82 & 11.55 & 19.94 & 10.70 & 10.89 & 8.20 \\
& DeepSeek-V3.2 & 9.56 & 9.56 & 8.24 & 11.35 & 12.33 & 8.10 & 11.71 & 8.00 \\
& GPT-5 Nano & 9.01 & 9.61 & 9.38 & 11.00 & 18.89 & 11.30 & 13.41 & 9.50 \\
\midrule
\textbf{RACE-Sched} & Qwen3-8B & 9.54 & 11.39 & \cellcolor{gray!20}5.33 & \cellcolor{gray!20}6.75 & \cellcolor{gray!20}6.43 & 4.80 & 11.09 & 9.50 \\
\textbf{(Ours)} & Qwen3-14B & 10.71 & 11.83 & 6.31 & 7.50 & 7.46 & 4.90 & \cellcolor{gray!20}9.37 & \cellcolor{gray!20}6.10 \\
& Qwen3-32B & 9.72 & 11.50 & 6.89 & 8.15 & 7.31 & \cellcolor{gray!20}4.50 & 10.67 & 7.30 \\
& DeepSeek-V3.2 & 8.02 & 9.22 & 7.35 & 9.30 & 7.87 & 5.40 & 11.01 & 7.90 \\
& GPT-5 Nano & 8.25 & 9.39 & 6.44 & 7.55 & 10.75 & 6.60 & 9.73 & \cellcolor{gray!20}6.10 \\
& Qwen3-Coder-30B & \cellcolor{gray!20}4.97 & \cellcolor{gray!20}6.50 & 8.89 & 11.65 & 11.23 & 7.80 & 13.42 & 10.50 \\
\bottomrule
\end{tabular}
\caption{Performance on GEN-Bench, MK-Bench, and JMS-Bench. Lower RPD and rank are better.}
\label{tab:main}
\end{table*}

\subsubsection{Benchmark Datasets and Evaluation Metrics}

We validate performance across three benchmark suites encompassing standardized testbeds to high-fidelity simulation settings. GEN-Bench assesses policy generalization capabilities, while MK-Bench extends classical Brandimarte instances to the dynamic domain~\cite{brandimarte1993routing}. JMS-Bench introduces industrial complexity through the simulation of semiconductor cluster tools with reentrant flows~\cite{cao2025reflecsched}. 
Regarding evaluation metrics, we measure solution quality using the Relative Percent Deviation (RPD) from the Best Known Solution and the average rank across all test instances. We further perform an analysis of cumulative token consumption and wall-clock time to assess the tradeoff between computational efficiency and real-time responsiveness.

\subsubsection{Baseline Methods and LLM Backends}

We evaluate RACE-Sched against a spectrum of symbolic, neural, and generative baselines. Classical approaches are exemplified by a GP dispatcher~\cite{mei2016feature}. Neural methods encompass the attention-based DAN~\cite{wang2023flexible} alongside three Reinforcement Learning agents: the breakdown robust IDDQN~\cite{wu2025dynamic}, the option-based PPO-OC~\cite{yuan2025deep}, and the hierarchical HMPSAC~\cite{ding2025data}. Finally, we compare against two LLM baselines: a zero-shot LLM-Direct agent and ReflecSched, which employs reflection-driven simulations~\cite{cao2025reflecsched}.

The proprietary GPT models include GPT-4o~\cite{openai2024gpt4ocard} and GPT-5 Nano~\cite{singh2025openaigpt5card}, while the open-weight models cover DeepSeek-V3~\cite{deepseekai2025deepseekv3technicalreport}, DeepSeek-V3.2~\cite{deepseekai2025deepseekv32pushingfrontieropen}, and the Qwen3 series with 8B, 14B, 32B parameters and Qwen3-Coder-30B~\cite{qwen3technicalreport}.
Inference is managed by using the vLLM~\cite{kwon2023efficient} high-throughput serving engine, hosted on a compute node configured with an Intel Xeon Platinum 8468V CPU and an NVIDIA H100 80GB GPU.

\subsection{Performance and Efficiency}

\subsubsection{Scheduling Performance Analysis}

As shown in Table~\ref{tab:main}, in normal scale scenarios, where combinatorial complexity is most pronounced, RACE-Sched exhibits superior convergence performance toward global optimality.
Specifically, the Qwen3-8b variant achieves an average RPD of $5.33\%$, outperforming the leading DRL baseline IDDQN with RPD $9.82\%$ by $45.7\%$.
This performance difference highlights the capacity of symbolic evolution to address long-horizon strategic dependencies, which are often obscured in the latent representations of neural approximations.

In Normal scale instances, RACE-Sched variants consistently occupy the upper tiers of the performance hierarchy.
Specifically, the Qwen3-8b configuration attains a mean rank of $6.75$, maintaining a significant lead over the top performing DRL baselines, IDDQN with rank $12.1$ and DAN with rank $12.55$.
In Small scale instances, despite the compressed optimization margins characteristic of smaller search spaces, RACE-Sched retains a distinct performance over LLM-Direct and ReflecSched.
This cross scale stability underscores that the synthesized symbolic code captures fundamental scheduling principles, enabling generalization that remains invariant to shifts in job volume and machine density.

\subsubsection{Token Consumption Assessment}

Table~\ref{tab:tokens} presents a comparative evaluation of cumulative token expenditure. The Semantic Knowledge Repository significantly reduces reliance on token-intensive generative exploration. Across all model backends ranging from Qwen3 to GPT-5, RACE-Sched consistently requires the lowest input and output volume. Specifically, under the Qwen3-8b configuration, our framework lowers input token consumption by approximately 60\% relative to LLM-Direct baseline. This efficiency gain arises from two structural advantages. First, the retrieval mechanism utilizes archived priors to prune the search space and eliminate redundant inferential steps. Second, the transition from per step action generation to compact policy synthesis drastically minimizes output volume. By confining LLM activation to performance critical triggers, the architecture effectively balances the complexity associated with deliberative reasoning and computational sustainability.

\begin{table*}[t]
\centering
\begin{tabular}{lrrrrrr}
\toprule
\textbf{Model Backend} & \multicolumn{3}{c}{\textbf{Mean Input Tokens}} & \multicolumn{3}{c}{\textbf{Mean Output Tokens}} \\
\cmidrule(lr){2-4} \cmidrule(lr){5-7}
& LLM-Direct & ReflecSched & RACE-Sched & LLM-Direct & ReflecSched & RACE-Sched \\
\midrule
Qwen3-8B & 59,977 & 69,386 & \cellcolor{gray!20}24,609 & 12,261 & 11,679 & \cellcolor{gray!20}4,802 \\
Qwen3-14B & 62,910 & 67,091 & \cellcolor{gray!20}47,581 & 9,423 & 18,725 & \cellcolor{gray!20}7,416 \\
Qwen3-32B & 78,832 & 68,905 & \cellcolor{gray!20}27,714 & 14,653 & 32,309 & \cellcolor{gray!20}3,900 \\
DeepSeek-V3 / V3.2 & 57,625 & 69,245 & \cellcolor{gray!20}44,565 & 14,277 & 8,488 & \cellcolor{gray!20}8,464 \\
GPT-4o & 58,009 & 64,031 & \cellcolor{gray!20}35,494 & 10,544 & 18,307 & \cellcolor{gray!20}4,675 \\
GPT-5 Nano & N/A & 70,435 & \cellcolor{gray!20}28,683 & N/A & 108,403 & \cellcolor{gray!20}50,353 \\
\bottomrule
\end{tabular}
\caption{Model-wise comparison of token consumption efficiency on GEN-Bench.}
\label{tab:tokens}
\end{table*}

\subsubsection{Real-Time Latency Evaluation}

Table~\ref{tab:latency} highlights the significant discrepancy between heuristic execution and deliberative reasoning. The Reactive Stream consistently exhibits sub-millisecond latency across all benchmark suites. Notably, the DeepSeek V3.2 and GPT-5 Nano configurations achieve a mean response time of less than 0.1 ms. This deterministic performance ensures that the tactical control loop complies with the strict real-time constraints imposed by industrial manufacturing environments.

In contrast, the Deliberative Stream functions on a timescale that is orders of magnitude longer with latencies ranging from 77 to 114 seconds per policy update. 
Synchronous integration of such high-latency reasoning would result in catastrophic blockages of the production line. RACE-Sched mitigates this risk by isolating the Deliberative Stream from the critical execution path. New rules are deployed only after validation via an atomic update. This architectural design maintains dispatching latency at the microsecond level, while enabling continuous long-horizon optimization.

\begin{table}[ht]
\centering
\small
\setlength{\tabcolsep}{3pt}
\renewcommand{\arraystretch}{1.05}
\begin{tabular}{p{0.26\columnwidth} l r r r}
\toprule
\textbf{Backend} & \textbf{Datasets} & \textbf{React ms} & \textbf{Delib s} & \textbf{Ratio $\times 10^6$} \\
\midrule
\multirow{3}{*}{DeepSeek-V3.2} & GEN-Bench & 0.054 & 77.05 & 1.43 \\
& MK-Bench & 0.084 & 113.79 & 1.35 \\
& JMS-Bench & 0.092 & 90.08 & 0.98 \\
\midrule
\multirow{3}{*}{GPT-5 Nano} & GEN-Bench & 0.050 & 114.08 & 2.28 \\
& MK-Bench & 0.060 & 111.94 & 1.87 \\
& JMS-Bench & 0.085 & 109.39 & 1.29 \\
\bottomrule
\end{tabular}
\caption{Reactive Stream latency and Deliberative update time across model backends.}
\label{tab:latency}
\end{table}

\subsection{Ablation Studies}

We compare four structural variants.
\begin{itemize}
    \item \textbf{Full}: The complete architecture integrating both repository retrieval and iterative deliberation for continuous rule optimization.
    \item \textbf{w/o Loop}: Disables the reasoning loop to generate a single candidate rule per step without retrospective feedback.
    \item \textbf{w/o Repo}: Removes knowledge retrieval to force a cold start initialization without historical priors.
    \item \textbf{w/o Both}: Excludes both components to serve as a baseline lacking memory and self correction mechanisms.
\end{itemize}

We retrieve archived rules based on normalized Manhattan distance within a meta-feature space defined by job and machine counts. To ensure reproducibility during benchmarking, a serialized execution protocol is implemented, while the deployed system maintains asynchronous decoupling.

\subsubsection{Contribution of Epistemic Transfer via Rule Repository}

Table~\ref{tab:ablation} highlights the critical role of the Semantic Knowledge Repository in mitigating cold start inefficiencies. The comparison between the Full architecture and the variant excluding the repository reveals a substantial performance gap. 
On the GEN-Bench Small dataset, the retrieval of  historical priors reduces the Relative Percent Deviation from 26.19\% to 10.41\%. This finding indicates that initializing deliberative reasoning with heuristics derived from similar problem instances effectively prunes the search space and minimizes reliance on stochastic exploration. 
Although the magnitude of improvement varies across specific datasets, the repository provides significant stability on a global scale. The Consolidated metrics confirm this structural advantage: the Full model achieves a global average deviation of 9.39\%, in contrast to 14.07\% for the unaugmented baseline.

\subsubsection{Impact of Recursive Agentic Refinement}

A comparison between the Full model and the variant without the loop reveals that refinement generally improves solution quality for complex tasks. For instance, on the GEN-Bench Normal dataset, enabling the loop reduces the RPD from 8.17\% to 6.16\%. This gain stems from the capability of the system to analyze sandbox feedback and rectify logical inconsistencies in subsequent iterations. 
Although the single-step approach performs competitively on simpler tasks, the Full architecture secures a superior consolidated average rank of 2.32 compared to 2.40 for the variant lacking the loop. This confirms that iterative validation provides greater stability across diverse problem distributions.

\begin{table}[t]
\centering
\footnotesize
\setlength{\tabcolsep}{2.2pt}
\renewcommand{\arraystretch}{1.08}
\begin{tabular}{@{}llrrrr@{}}
\toprule
\textbf{Dataset} & \textbf{Metric} & \textbf{Full} & \textbf{No Repo} & \textbf{No Loop} & \textbf{No Both} \\
\midrule
\multirow{2}{*}{GEN-S} & RPD (\%) & 10.41 & 26.19 & 10.18 & 31.02 \\
& Rank & 1.94 & 2.61 & 1.78 & 3.00 \\
\midrule
\multirow{2}{*}{GEN-N} & RPD (\%) & 6.16 & 9.54 & 8.17 & 13.77 \\
& Rank & 2.15 & 3.25 & 2.50 & 3.60 \\
\midrule
\multirow{2}{*}{MK} & RPD (\%) & 11.78 & 9.43 & 14.16 & 16.21 \\
& Rank & 2.20 & 2.40 & 2.60 & 3.20 \\
\midrule
\multirow{2}{*}{JMS} & RPD (\%) & 9.22 & 11.10 & 8.52 & 14.80 \\
& Rank & 3.00 & 2.70 & 2.70 & 2.90 \\
\midrule
Global & Avg. RPD (\%) & \cellcolor{gray!20}9.39 & 14.07 & 10.26 & 18.95 \\
Consol. & Avg. Rank & \cellcolor{gray!20}2.32 & 2.74 & 2.40 & 3.18 \\
\bottomrule
\end{tabular}
\caption{Ablation results quantifying the contribution of the Repository and Iterative Loop. GEN-S and GEN-N denote the Small and Normal subsets of GEN-Bench, respectively.}
\label{tab:ablation}
\end{table}

\subsection{Resilience Under Machine Failures}

\begin{figure}[!htbp]
    \centering
    \includegraphics[width=0.48\textwidth]{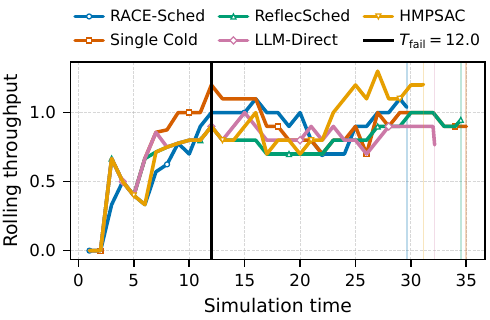}
    \caption{Rolling throughput after a machine failure.}
    \label{fig:rolling_throughput}
\end{figure}

We evaluate operational robustness under nonstationary conditions using a stress test within the JMS Bench framework. Figure~\ref{fig:rolling_throughput} depicts the system's response to an abrupt capacity loss at a critical workstation, which occurs at time $T_{fail} = 12.0$.

The baseline approaches demonstrate inadequate adaptability to this structural shift. The Single Cold variant results in the worst makespan of 35.0 due to its reliance on a static heuristic that cannot adjust to the new bottleneck. Correspondingly, the DRL baseline HMPSAC exhibits high variance and a prolonged recovery period, which suggests insufficient generalization to out-of-distribution states.

In contrast, RACE-Sched demonstrates superior resilience. Upon detection of the throughput degradation, the Deliberative Stream asynchronously synthesizes and deploys a new priority rule tailored to the altered system dynamics. This rapid adaptive response restores throughput to pre-failure levels and results in the minimum final makespan of 29.6. These results confirm that the asynchronous decoupling effectively supports continuous policy refinement without blocking the real-time control loop.

\section{Conclusion}

We introduce RACE-Sched to reconcile the reasoning capability of LLMs with the latency constraints of industrial control systems. By decoupling fast heuristic execution from asynchronous policy synthesis, RACE-Sched maintains millisecond-level responsiveness while continuously evolving symbolic control logic. The Semantic Knowledge Repository reduces redundant deliberation through retrieval-based reuse, and sandbox validation prevents unverified rules from entering the online control loop. Extensive experiments across GEN-Bench, MK-Bench, and JMS-Bench show that RACE-Sched outperforms state-of-the-art DRL and generative scheduling baselines while preserving low-latency online execution.

RACE-Sched also has several deployment limitations. Its rule evolution quality depends on the capability of the background LLM, since weaker models may generate invalid or sub-optimal candidate rules that are rejected by sandbox validation. The construction of representative sandbox scenarios and trigger conditions requires engineering effort, because the validation environment must reflect the main disturbance patterns of the target shop floor. Although the asynchronous design prevents deliberation from blocking the control loop, reaction to brand-new disturbance types can still be delayed when candidate synthesis and validation require multiple deliberation cycles.

\section*{Ethical Statement}

There are no ethical issues.

\section*{Acknowledgments}

This work was partially supported by Shenzhen Loop Area Institute under Project No. FP202602. The authors gratefully acknowledge the support from the project team members involved in this work.

\bibliographystyle{named}
\bibliography{ijcai26}

\end{document}